\documentclass[]{spie}  

\usepackage{amsmath,amsfonts,amssymb}
\usepackage{graphicx}
\usepackage[colorlinks=true, allcolors=blue]{hyperref}

\usepackage{array, booktabs}
\usepackage{float}  
\usepackage{placeins} 
\usepackage{comment}
\usepackage{csquotes}
\usepackage{subcaption}
\usepackage{xcolor}

\def\equationautorefname~#1\null{Eq.~(#1)\null}

\title{Toward Non-Invasive Diagnosis of Bankart Lesions with Deep Learning}

\author[1]{Sahil Sethi}
\author[1]{Sai Reddy}
\author[2]{Mansi Sakarvadia}
\author[3]{Jordan Serotte}
\author[3]{Darlington Nwaudo}
\author[3]{Nicholas Maassen}
\author[3]{Lewis Shi}
\affil[1]{Pritzker School of Medicine, University of Chicago, 
IL, USA
}
\affil[2]{Department of Computer Science, University of Chicago, 
IL, 
USA}
\affil[3]{Department of Orthopaedic Surgery \& Rehabilitation Medicine,
UChicago Medicine, 
IL, USA}

\authorinfo{Author information: (Send correspondence to S.S.)\\S.S.: E-mail: sethis@uchicago.edu\\  L.S.: E-mail: lshi@bsd.uchicago.edu}

\begin{document} 
\maketitle

\begin{abstract}

\noindent \textbf{Purpose:} Bankart lesions, or anterior-inferior glenoid labral tears, are diagnostically challenging on standard MRIs due to their subtle imaging features—often necessitating invasive MRI arthrograms (MRAs). This study develops deep learning (DL) models to detect Bankart lesions on both standard MRIs and MRAs, aiming to improve diagnostic accuracy and reduce reliance on MRAs.

\noindent \textbf{Methods:} We curated a dataset of 586 shoulder MRIs (335 standard, 251 MRAs) from 558 patients who underwent arthroscopy. Ground truth labels were derived from intraoperative findings, the gold standard for Bankart lesion diagnosis. Separate DL models for MRAs and standard MRIs were trained using the Swin Transformer architecture, pre-trained on a public knee MRI dataset. Predictions from sagittal, axial, and coronal views were ensembled to optimize performance. The models were evaluated on a 20\% hold-out test set (117 MRIs: 46 MRAs, 71 standard MRIs).

\noindent \textbf{Results:} Bankart lesions were identified in 31.9\% of MRAs and 8.6\% of standard MRIs. The models achieved AUCs of 0.87 (86\% accuracy, 83\% sensitivity, 86\% specificity) and 0.90 (85\% accuracy, 82\% sensitivity, 86\% specificity) on standard MRIs and MRAs, respectively. These results match or surpass radiologist performance on our dataset and reported literature metrics. Notably, our model's performance on non-invasive standard MRIs matched or surpassed the radiologists interpreting MRAs.

\noindent \textbf{Conclusion:} This study demonstrates the feasibility of using DL to address the diagnostic challenges posed by subtle pathologies like Bankart lesions. Our models demonstrate potential to improve diagnostic confidence, reduce reliance on invasive imaging, and enhance accessibility to care.

\end{abstract}


\noindent \textit{Accepted for presentation at SPIE Medical Imaging 2025: Computer-Aided Diagnosis. The manuscript is expected to appear in the conference proceedings.}

\section{INTRODUCTION}
\label{sec:intro}  
Glenoid labral tears are among the most common injuries to the glenohumeral joint, often presenting with pain, instability, decreased range of motion, and a popping/locking sensation in the shoulder \cite{guanche_clinical_2003, keener_superior_2009}. These lesions can be caused by trauma, shoulder instability/dislocation, or overuse such as through overhead throwing activities \cite{liu_diagnosis_1996}. Annually, it is estimated that 6\% of the general population and 35\% of the sporting population in the United States are impacted by glenoid labral tears, with the incidence increasing every year \cite{zughaib_outcomes_2017}. Among the glenoid labral tear variants, the anterior-inferior subtype, also known as a Bankart lesion, is 
common in patients who dislocate their shoulder—with an incidence of 59\% in first-time shoulder dislocations and 66\% in recurrent dislocations \cite{rutgers_recurrence_2022}. \par

Without accurate diagnosis and timely treatment, patients commonly experience exacerbation of functional impairments, pain, muscle weakness, and chronic shoulder instability leading to recurrent subluxation and eventual bone loss \cite{kang_complications_2009, rutgers_recurrence_2022}. These ramifications can cause significant patient burden and societal costs, and are exacerbated when glenoid labral tears are misdiagnosed due to similar clinical presentations with other conditions including arthritis, rotator cuff disorders, and tendonitis \cite{keener_superior_2009}. \par

The gold standard for diagnosing Bankart lesions is arthroscopy: direct visualization of the labrum with a camera during minimally invasive surgery \cite{rutgers_recurrence_2022}. Previous studies evaluating the performance of diagnostic tools for labral tears have used this as the ground truth \cite{major_evaluation_2011, ni_deep_2022, zlatkin_assessment_2004}. However, this is a surgical procedure—often requiring pre-operative clearance and accompanied by the risks of anesthesia—and thus is usually performed for repair rather than diagnosis of the lesion \cite{mazzocca_traumatic_2011, chang_ssr_2024}. \par

\begin{figure}[t]
   \begin{center}
   \begin{tabular}{c} 
   \includegraphics[height=7.5cm]{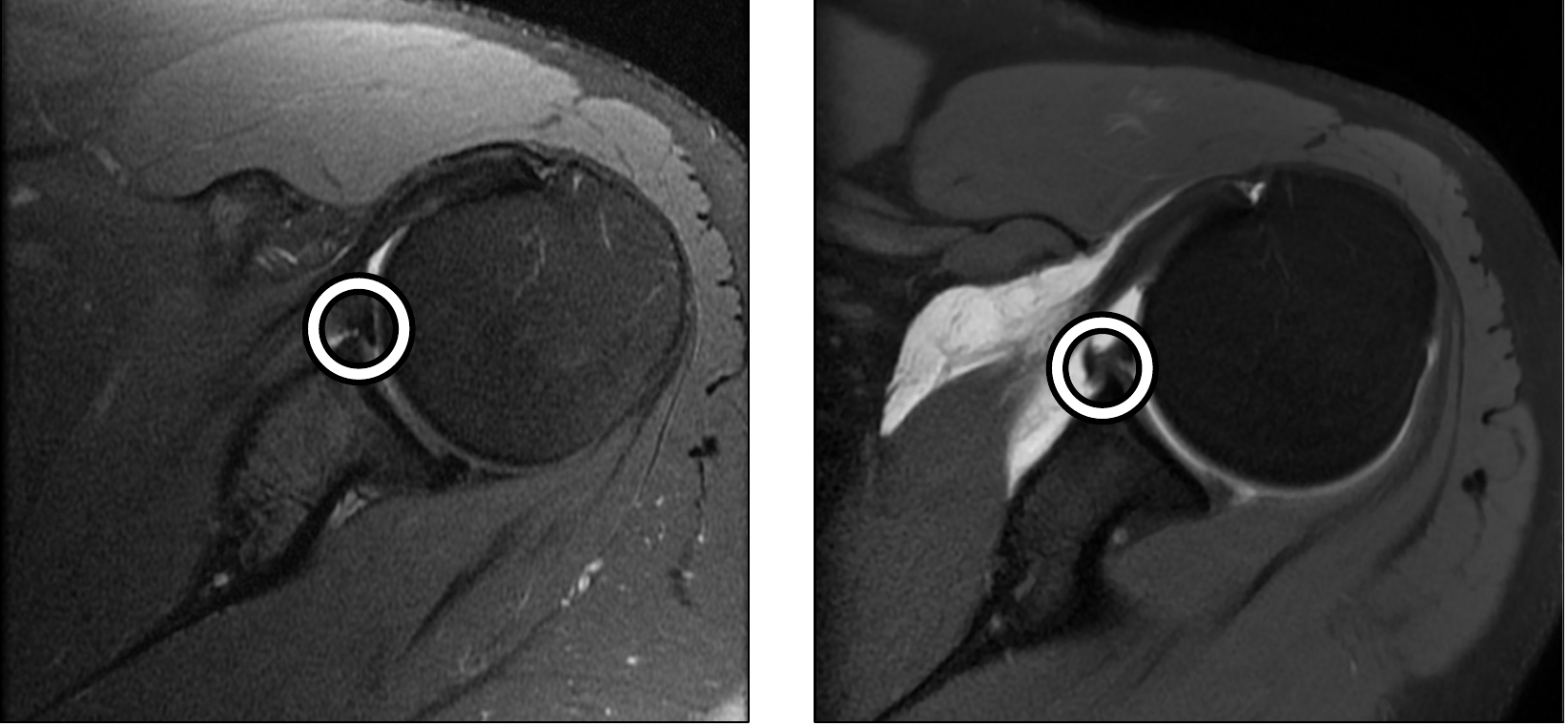}
   \end{tabular}
   \end{center}
   \caption[example] 
   { \label{fig:introfig} 
Bankart lesion on standard MRI (left) and MRI arthrogram (right) in the axial view. Images are from the same patient and depict the same tear. White circles reflect annotations identifying the tear,  provided by a shoulder/elbow fellowship-trained orthopedic surgeon.}
   \end{figure}

MRI arthrograms (MRAs) are generally the preferred imaging modality for pre-operative diagnosis of glenoid labral tears, including Bankart lesions \cite{chandnani_glenoid_1993}, though they are less accurate than direct visualization via arthroscopy \cite{rutgers_recurrence_2022}. MRAs are performed by injecting contrast into the joint with a large needle prior to the scan, which improves the delineation of intra-articular structures and signal-to-noise ratio compared to non-contrast (standard) MRIs \cite{chang_ssr_2024}. The improved visualization of intra-articular structures is apparent in \autoref{fig:introfig}, which depicts the same Bankart lesion on an MRA and a standard (non-contrast) MRI. 

MRA sensitivity ranges from 74-96\% and specificity from 91-98\% \cite{rixey_accuracy_2023, magee_3-t_2009, woertler_mr_2006} in detecting glenoid labral tears. Despite the diagnostic capabilities of MRAs, the use of intra-articular contrast increases the time and cost of performing the imaging \cite{liu_comparison_2020}. Intra-articular contrast injection is generally considered safe, but roughly two-thirds of 
patients experience delayed-onset pain in the affected area \cite{chang_ssr_2024, giaconi_morbidity_2011, ali_radio-carpal_2022}. 
There is also a risk of allergic reaction (0.4\% for hives and 0.003\% for severe anaphylaxis), vasovagal reactions (1.4\%), and a joint infection (0.003\%) \cite{chang_ssr_2024, newberg_complications_1985, hugo_complications_1998}.

While the standard (non-contrast) MRI is a non-invasive alternative to arthroscopy and MRAs, the lack of intra-articular contrast makes it more difficult to visualize changes to the labrum. Consequently, radiologists generally have poorer performance in glenoid labral tear diagnosis using standard MRIs compared to MRAs, with some studies reporting sensitivity ranging from $52\%-55\%$ and specificity ranging from $89\%-100\%$ \cite{chandnani_glenoid_1993, zlatkin_assessment_2004, magee_sensitivity_2006,
arnold_non-contrast_2012, alexeev_variability_2021}. 

To address these limitations, this study develops and evaluates deep learning (DL) models for detecting Bankart lesions on both MRAs and standard MRIs. While DL has shown promise in other areas of medicine, including pneumonia detection and diabetic retinopathy grading \cite{sun_lesion-aware_2021, calli_deep_2021}, its application in orthopedic imaging remains underexplored. By leveraging an ensemble approach across sagittal, axial, and coronal views, our models achieve comparable sensitivity and specificity on standard MRIs compared to radiologists on invasive MRAs, demonstrating the potential of DL to address this diagnostic challenge. 

\section{Data}
\label{sec:data}
\subsection{Dataset Collection}

\begin{figure}[t]
   \begin{center}
   \begin{tabular}{c} 
   \includegraphics[width=\textwidth]{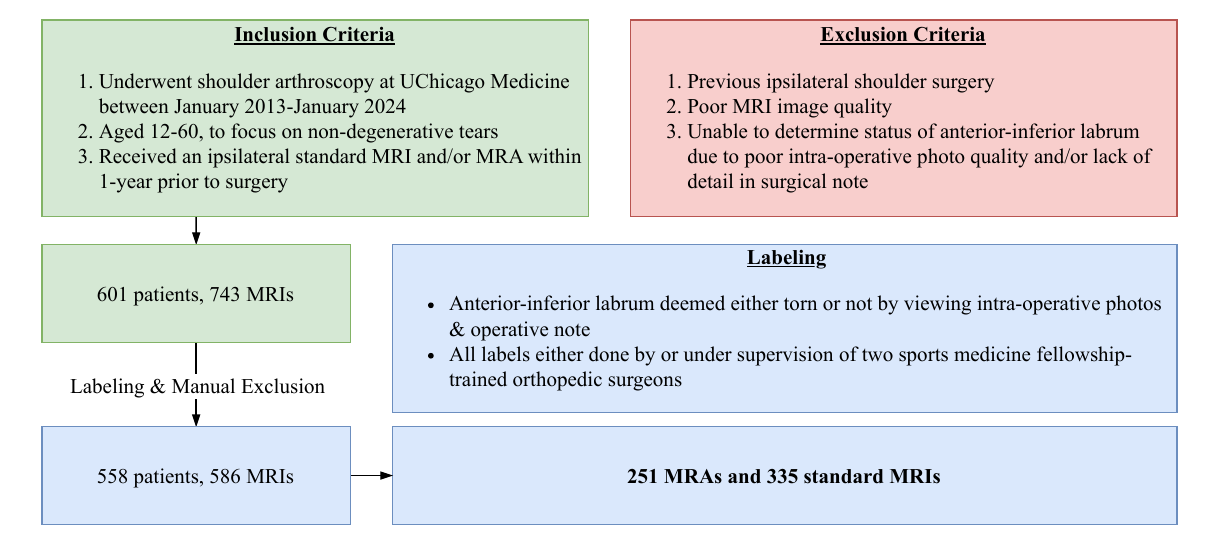}
   \end{tabular}
   \end{center}
   \caption[example] 
   { \label{fig:data_collection} 
Data Collection and Labeling Protocol.}
\end{figure}

\autoref{fig:data_collection} overviews the data collection and labeling protocol. We curated a dataset of patients who underwent shoulder arthroscopy at our institution between January 2013 and January 2024. Patients aged 12 to 60 who received a standard MRI and/or MRA within one year prior to arthroscopy were included. Younger patients were excluded to avoid age-related pathology differences, and older patients were excluded to focus on acute, operative labral tears rather than degenerative tears.

After excluding patients with prior ipsilateral surgery, incomplete imaging, or insufficient intraoperative documentation, 546 patients with 586 MRIs (335 standard, 251 MRA) were retained. Tear labels were derived from intraoperative arthroscopy findings, the diagnostic gold standard. Labels were curated by two shoulder/elbow fellowship-trained orthopedic surgeons and two orthopedic surgery residents trained by the surgeons. A common subset of 20 MRIs was labeled by all raters to measure inter-rater reliability—they achieved a Fleiss’s kappa=1.0, indicating complete agreement.

\subsection{Dataset Statistics \& Characteristics}
The final dataset consisted of 558 patients and 586 MRIs (some patients had bilateral surgeries). Demographics and clinical characteristics are summarized in \autoref{tab:demographics} and align with current clinical practices. MRA patients were younger and had a higher prevalence of labral tears (31.9\% vs. 8.6\% for standard MRIs; p $<$ 0.001). These differences reflect that MRAs are often reserved for patients with higher clinical suspicion of pathology \cite{woertler_mr_2006}. Further dataset details are provided in \autoref{app:dataset}. Younger patients, who are more likely to be suspected of having labral tears, are often preferentially given MRAs due to their demographic and activity levels.

\subsection{Image Preprocessing}
MRIs were preprocessed to prepare them for deep learning analysis. Volumes were resized to $n$ x 400 x 400 pixels and center-cropped to 224 x 224 to isolate the region of interest. Intensity values were standardized by sequence type (e.g., T1, T2, PD) and fat saturation status using distribution statistics from the training set. Intensities were then scaled so that the voxel values ranged between 0 and 1. To address class imbalance, ten-fold augmentation of training samples was performed using random rotations, translations, scaling, flips, and Gaussian noise.

\begin{table}[H]
\centering
\caption{Demographics and Dataset Characteristics}
\renewcommand{\arraystretch}{1.2}  
\newcolumntype{P}[1]{>{\centering\arraybackslash}p{#1}}
\begin{tabular}{p{0.35\linewidth}  P{0.12\linewidth}  P{0.12\linewidth}  P{0.12\linewidth} P{0.12\linewidth}}
\toprule
\textbf{Characteristic} & \textbf{Total MRIs\textsuperscript{1}} & \textbf{MRI Arthrograms} & \textbf{Non-Enhanced MRIs} & \textbf{p-value\textsuperscript{2}}  \\
\midrule
\textbf{} & \textbf{n = 586} & \textbf{n = 251} & \textbf{n = 335} & \\ 
\midrule
\textbf{Number of Patients\textsuperscript{3}} & 546 & 238 & 318 & - \\ 
\textbf{Average Age (SD)} & 41 (13.9) & 31 (12.7) & 48.6 (9.3) & p $<$ 0.001 \\ 
\textbf{Female Sex (\%)} & 256 (43.7) & 88 (35.1) & 168 (50.1) & p $<$ 0.001\\ 
\textbf{Right-Sided Exams (\%)} & 365 (62.3) & 153 (61.0) & 212 (63.3) & p = 0.565\\ 
\textbf{3.0T Exams (\%)} & 367 (62.6) & 184 (73.3) & 183 (54.6) & p $<$ 0.001\\ 
\midrule
\textbf{\# MRIs with Bankart Lesions (\%)} & 109 (18.5) & 80 (31.9) & 29 (8.6) & p $<$ 0.001\\ 
\midrule
\textbf{Dataset Split} & & & & \\ 
Training MRIs (\% with tear) & 410 (18.5) & 186 (30.1) & 224 (8.7) & - \\ 
Validation MRIs (\% with tear) & 59 (18.6) & 19 (36.8) & 40 (10.0) & - \\ 
Testing MRIs (\% with tear) & 117 (18.8) & 46 (37.0) & 71 (7.0) & - \\ 
\bottomrule
\end{tabular}
\vspace{5pt} 
\parbox{0.95\textwidth}{%
\footnotesize%
\textsuperscript{1} All available sequences were included for each MRI (e.g., T1, T2, MERGE, PD, STIR), yielding 1109 axial, 1647 coronal, 978 sagittal, and 237 ABER (Abduction and External Rotation) sequences; ABER sequences were excluded due to insufficient numbers for model training. \\
\textsuperscript{2} p-values obtained via chi-squared for categorical variables and unpaired two-tailed t-test for continuous variables. \\
\textsuperscript{3} Total does not add up as some patients had surgeries on both sides, and may have received an MRA on one side and a standard MRI on the other. \\

}
\label{tab:demographics}
\end{table}

\section{Methods}
\label{sec:methods}
This study aimed to assess the feasibility of detecting Bankart lesions using deep learning models on MRI data. To address this challenge, we pre-trained a Swin Transformer V1\cite{liu_swin_2021} model on the MRNet knee MRI dataset \cite{bien_deep-learning-assisted_2018}, then fine-tuned on our dataset separately for MRAs and standard MRIs. Pre-training on MRNet, using the \enquote{abnormal} label, allowed the model to leverage representations learned from a related medical imaging task.

\subsection{Model Architecture} 
\label{model_architecture}
\begin{figure} [ht]
\begin{center}
\begin{tabular}{c} 
   \includegraphics[width=\textwidth]{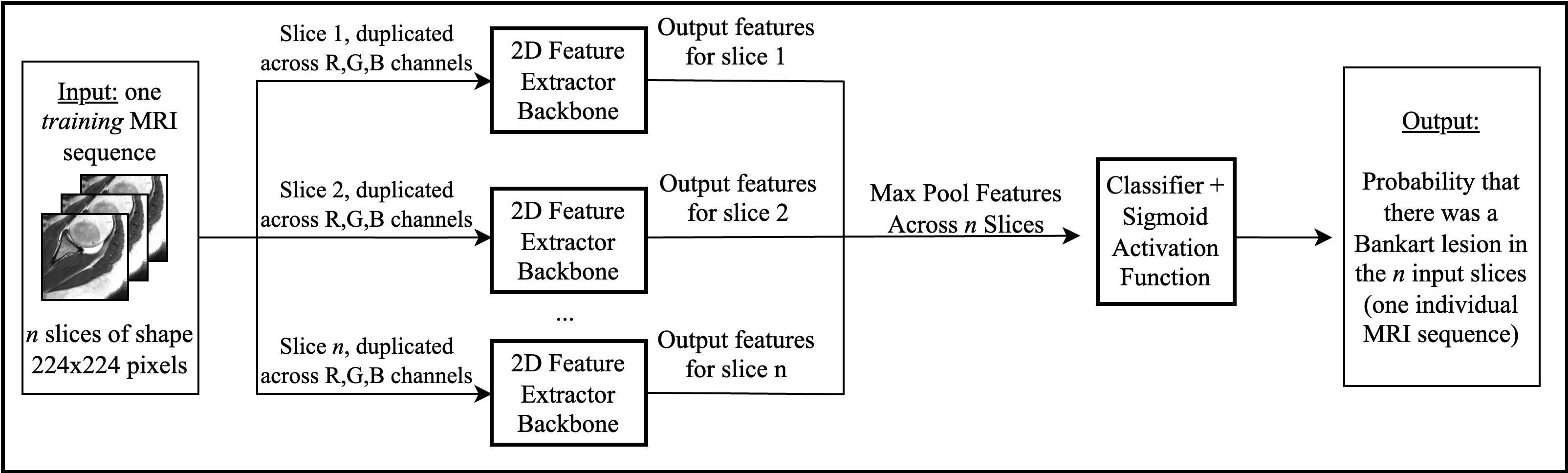} \\  
   \textbf{(a)} Schematic of 2D model training setup using 3D MRIs \\  
   \includegraphics[width=\textwidth]{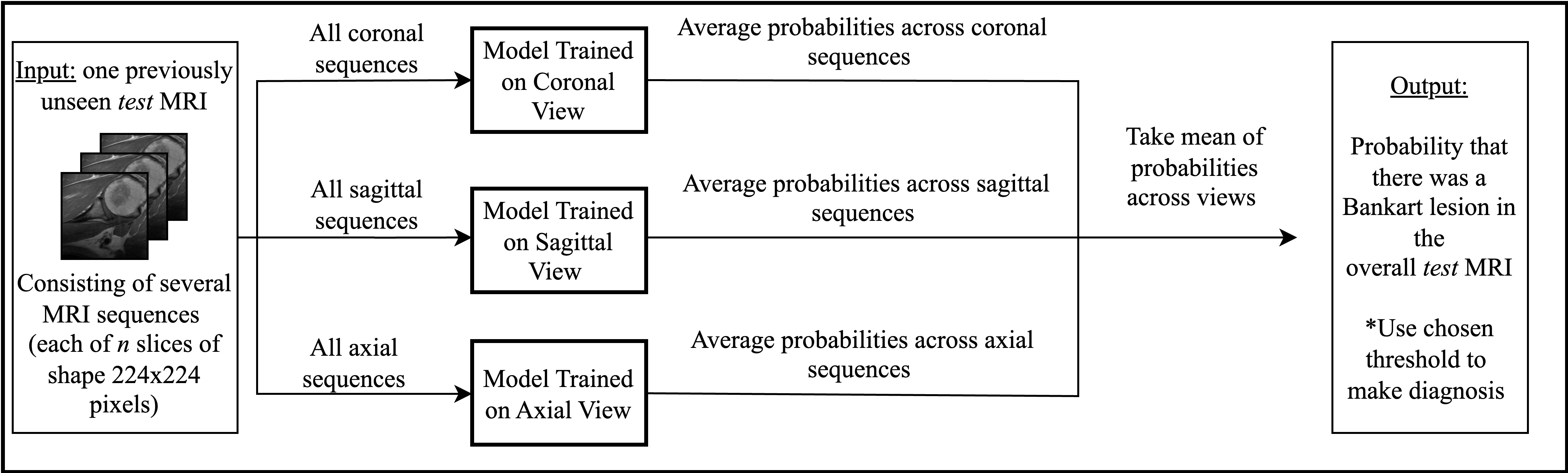} \\   
   \textbf{(b)} Schematic of model inference \\ 
\end{tabular}
\end{center}
\caption[Model Training \& Inference]
{\label{fig:train_inf} Model Training \& Inference
(a) Schematic of 2D model training setup using 3D MRIs. (b) Schematic of model inference. }
\end{figure}

Each MRI was processed as a series of 2D slices, which were passed through a Swin Transformer\cite{liu_swin_2021} feature extractor. For the pre-training step, ImageNet \cite{deng_imagenet_2009} weights were used for initialization. Slice-level features were aggregated using max pooling to produce a vector representing the entire scan. This vector was passed through a classifier layer to output a per-scan probability. Probabilities from sagittal, axial, and coronal models were averaged together to produce multi-view predictions. The prediction threshold used for accuracy, sensitivity, and specificity was set using the threshold at which sensitivity and specificity were equal for the multi-view ensemble on the validation set. This method was chosen to balance avoiding missed diagnoses while minimizing unnecessary interventions. The training and inference pipeline, which integrates these steps, is illustrated in \autoref{fig:train_inf}.

\subsection{Training and Inference} The Swin Transformer\cite{liu_swin_2021} model was trained using binary cross-entropy loss, scaled to account for class imbalance, ensuring equal contribution from underrepresented classes during optimization. During training, early stopping was employed based on validation accuracy, with a patience of 10 epochs, and model weights from the epoch with the highest validation accuracy were selected for inference.

The final model was evaluated on a hold-out test set (20\% of the dataset) using multi-view ensembling. Performance metrics, including accuracy, sensitivity, specificity, and AUC-ROC, were calculated to assess diagnostic performance compared to the original radiology reports from our dataset (with intraoperative findings as the ground truth). 

\section{Results \& Discussion}
\label{sec:results_discuss}
\subsection{Single-View \& Multi-View Ensemble ROC Performance}

\begin{figure*}[t]
\centering

    \begin{subfigure}[b]{0.45\textwidth}
        \centering
        \includegraphics[width=\textwidth]
        {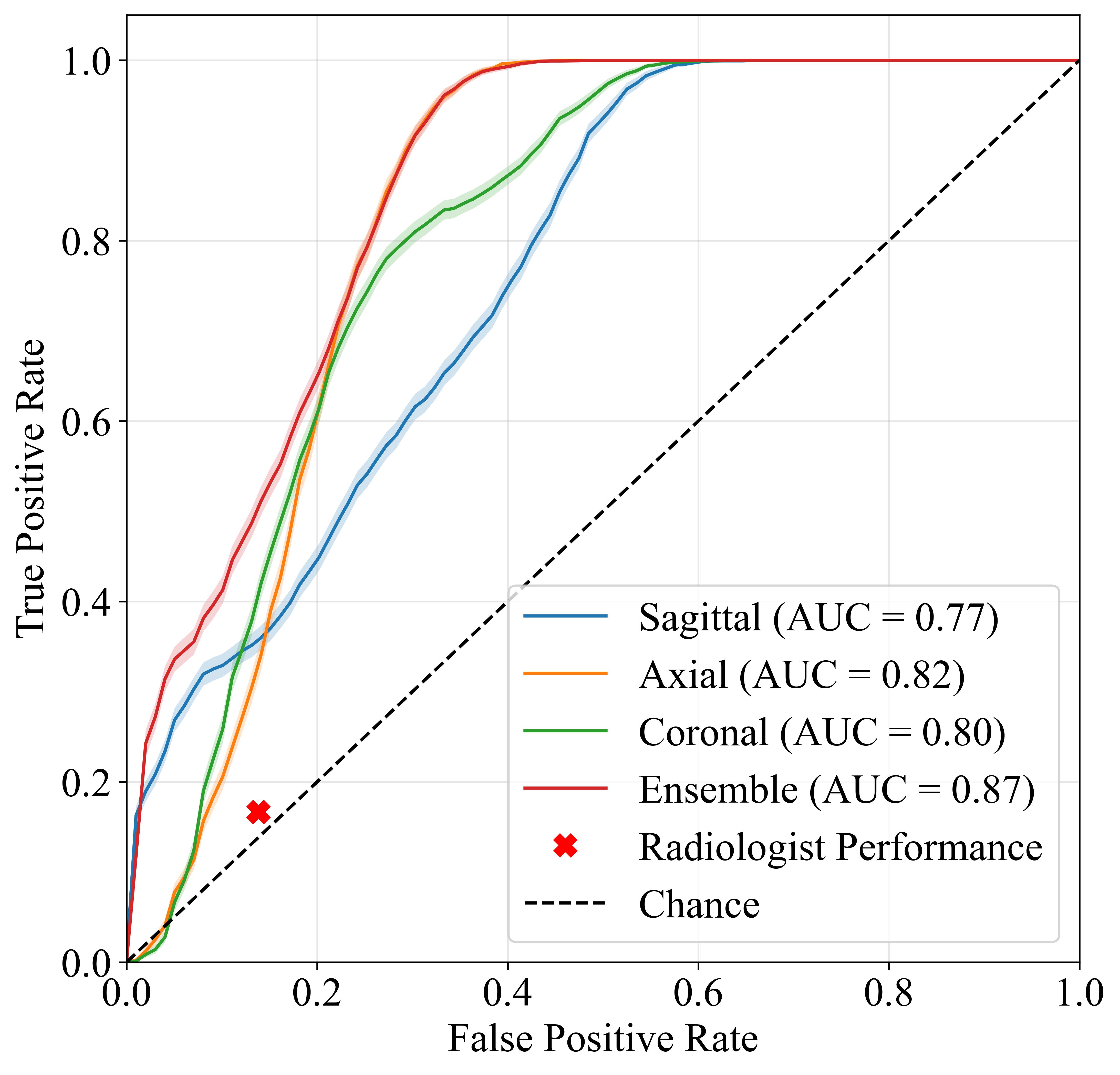}
        \caption{\textbf{MRI}}
        \label{fig:roc_wo}
    \end{subfigure}%
    \hfill
    \begin{subfigure}[b]{0.45\textwidth}
        \centering
        \includegraphics[width=\textwidth]
        {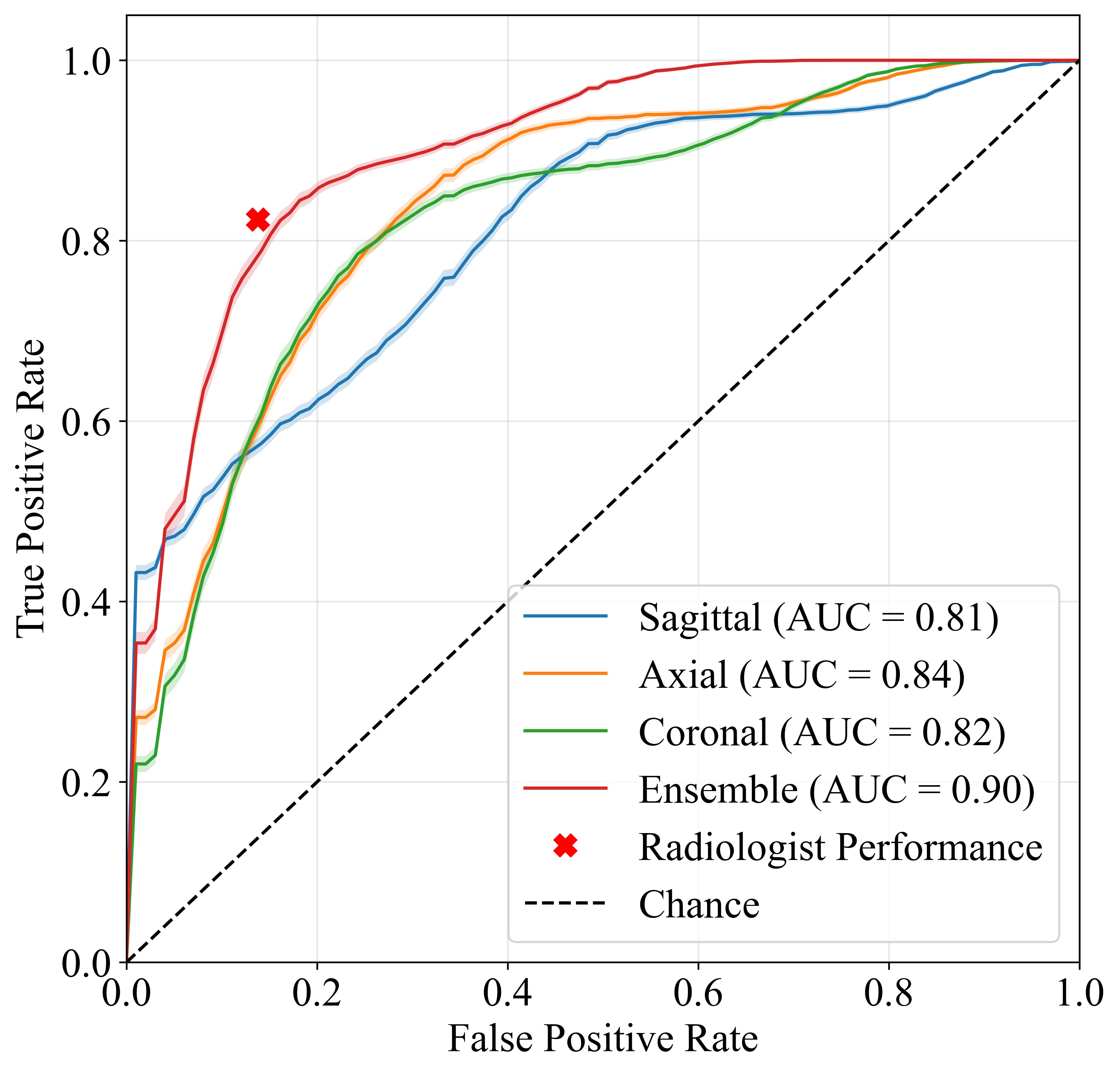}
        \caption{\textbf{MRA}}
        \label{fig:roc_w}
    \end{subfigure}%

    \caption{ \label{fig:ensemble_results} 
Receiver operating characteristic (ROC) curves for single-view models and the multi-view ensemble, compared to radiologist performance on (a) standard MRIs and (b) MRAs. The single-view models correspond to those included in the multi-view ensemble. Shaded regions around each curve represent 95\% confidence intervals, calculated through bootstrapping with 1000 iterations. Radiologist performance is marked with red \textbf{\textcolor{red}{x}} symbols, illustrating sensitivity and false positive rates derived from original radiology reports. The dashed diagonal line indicates the performance of a random classifier (AUC = 0.50).
    }
    
\end{figure*}

\noindent \textbf{Result:} The receiver operating characteristic (ROC) curves in \autoref{fig:ensemble_results} depict the performance of the single-view models and the multi-view ensemble on the hold-out test set.  

\noindent \textbf{Discussion:} The multi-view ensemble demonstrated superior diagnostic accuracy compared to single-view models on the hold-out test set, as shown in \autoref{fig:ensemble_results}. For standard MRIs, the ensemble achieved an AUC of 0.87, surpassing the single-view models for sagittal, axial, and coronal views (AUCs: 0.77, 0.82, and 0.80, respectively). Similarly, for MRAs, the ensemble achieved an AUC of 0.90, outperforming single-view models (AUCs: 0.81 for sagittal, 0.84 for axial, and 0.82 for coronal). These results demonstrate the ensemble’s ability to leverage complementary features across views to enhance diagnostic performance.

\subsection{Sensitivity, Specificity \& Accuracy (Threshold-Dependent Metrics)}

\noindent \textbf{Results: } For the model to have predicted a Bankart lesion diagnosis, the multi-view ensemble's output probability must have exceeded the thresholds of 0.71 and 0.19 for standard MRIs and MRAs, respectively. These thresholds were chosen as described in \autoref{model_architecture}. Results of the multi-view ensemble on the hold-out test set using these thresholds are depicted in \autoref{tab:model_performance}. 

On standard MRIs, the ensemble model achieved 85.9\% accuracy and 83.3\% sensitivity, significantly exceeding radiologist sensitivity from the original radiology reports (16.7\%) and matching their specificity (86.2\%). However, this specificity metric falls slightly outside the literature range of 89-100\% \cite{zlatkin_assessment_2004, arnold_non-contrast_2012}. 

For MRAs, the ensemble achieved 82.4\% sensitivity, matching both radiologists on our dataset and literature-reported values, although its specificity (86.2\%) was slightly lower than literature ranges for radiologists on MRAs (91-98\%)  \cite{rixey_accuracy_2023, magee_3-t_2009, woertler_mr_2006}. 

\begin{table}[t]
\centering
\caption{Final Ensemble Model Performance on Hold-Out Test Set}
\renewcommand{\arraystretch}{1.2}  
\newcolumntype{P}[1]{>{\centering\arraybackslash}p{#1}}
\begin{tabular}{p{0.3\linewidth} P{0.15\linewidth} P{0.15\linewidth} P{0.15\linewidth} P{0.1\linewidth}}
\toprule
\textbf{} & \textbf{Accuracy} & \textbf{Sensitivity (Recall)} & \textbf{Specificity} & \textbf{AUC-ROC\textsuperscript{1}} \\
\midrule
\textbf{Standard MRIs} (n=71) & & & & \\ 
\emph{Model} & 85.92\% (61/71)& 83.33\% (5/6)& 86.15\% (56/65)& 0.8718 \\ 
\emph{Radiology Reports} & 80.28\% (57/71)& 16.67\% (1/6)& 86.15\% (56/65)&  -  \\ 
\emph{Literature Radiologists\textsuperscript{2}} & - & 52-55\% & 89-100\% & - \\ 

\midrule
\textbf{MRI Arthrograms} (n=46) & & & & \\ 
\emph{Model} & 84.78\% (39/46)& 82.35\% (14/17)& 86.21\% (25/29)& 0.9006 \\ 
\emph{Radiology Reports} & 84.78\% (39/46)& 82.35\% (14/17)& 86.21\% (25/29)& -  \\ 
\emph{Literature Radiologists\textsuperscript{2}} & - & 74-96\% & 91-98\% & - \\ 
\bottomrule
\end{tabular}
\vspace{5pt} 
\parbox{0.95\textwidth}{%
\footnotesize%
\textsuperscript{1} Area Under the Receiver Operating Curve (AUC-ROC)  \\
\textsuperscript{2} Values obtained for glenoid labral tears in general from largest studies available in the literature for standard MRIs  \cite{zlatkin_assessment_2004, arnold_non-contrast_2012} and MRAs \cite{rixey_accuracy_2023, magee_3-t_2009, woertler_mr_2006}. \\
}
\label{tab:model_performance}
\end{table}

\noindent \textbf{Discussion:} The ensemble’s performance on standard MRIs is particularly noteworthy, as it surpasses radiologist accuracy and sensitivity on our dataset using both standard MRIs and MRAs (see \autoref{tab:model_performance}), and aligns with sensitivity ranges reported in the literature for radiologists on MRAs \cite{rixey_accuracy_2023, magee_3-t_2009, woertler_mr_2006}. The slightly increased false-positive rate (lower specificity) of our model on both standard MRIs and MRAs compared to literature metrics is unlikely to have significant clinical implications, as imaging results are interpreted in the context of patient history and physical examination \cite{liu_diagnosis_1996, walsworth_reliability_2008}. Patients undergoing imaging often present with significant pain or a strong clinical suspicion of a tear \cite{liu_diagnosis_1996, walsworth_reliability_2008}. In these situations, false positives are unlikely to result in unnecessary treatments but may instead prompt more careful clinical follow-up or additional diagnostic tests \cite{rutgers_recurrence_2022}. Importantly, the ensemble’s high sensitivity ensures that tears are rarely missed, a critical factor for timely and effective management. This improvement addresses a critical limitation of standard MRIs in detecting Bankart lesions, where missed tears often necessitate invasive MRAs. 

\subsection{Clinical Implications:} The ensemble’s ability to achieve comparable or superior diagnostic performance to radiologists on the MRAs with non-invasive standard MRIs underscores its potential to transform clinical workflows. By reducing the reliance on MRAs, this approach could decrease patient burden, avoid unnecessary procedures, and lower healthcare costs. Additionally, improving diagnostic accuracy on standard MRIs could streamline patient care in resource-limited settings where access to MRAs may be restricted. However, these results are preliminary and were achieved on a single-center dataset. Future work will be critical to validate these findings on larger, multi-institutional datasets and assess their generalizability to broader clinical populations.

\section{Related Work}
 This study presents the first application of deep learning (DL) to detect anterior-inferior glenoid labrum tears (Bankart lesions). To our knowledge, only two prior studies have applied DL to glenoid labrum pathology, and both focused exclusively on superior labrum anterior-to-posterior (SLAP) tears. Ni et al. \cite{ni_deep_2022} evaluated a custom CNN-based architecture on MRI arthrograms, while Clymer et al. \cite{clymer_applying_2019} explored self-supervised pre-training to improve performance on a very small dataset (34 standard MRIs). Both studies limited their models to specific MRI sequences—Ni et al. used only axial and oblique-coronal fat-saturation T1-weighted fast spin‒echo sequences, and Clymer et al. focused on T2 fat-saturation coronal sequences.

 In contrast, our study incorporates all available MRI sequences for each view (e.g., T1, T2, MERGE, PD, STIR), ensembles multiple views, and applies DL to both standard MRIs and MRI arthrograms. These advancements aim to overcome prior limitations and address the broader clinical challenge of diagnosing Bankart lesions, a pathology with distinct imaging and diagnostic difficulties.


\section{Conclusions}
We demonstrate the feasibility of using deep learning (DL) to detect Bankart lesions on both standard MRIs and MRI arthrograms (MRAs), achieving high diagnostic accuracy on both modalities. Importantly, our ensemble achieved diagnostic performance on non-invasive standard MRIs that rivaled radiologists interpreting invasive MRAs in our dataset. This suggests the potential to reduce reliance on arthrograms, benefiting patients by avoiding invasive procedures, lowering healthcare costs, and improving diagnostic accessibility.

While these results are promising, they represent a single-center study with a relatively small dataset. Future work should prioritize external validation on larger, multi-institutional datasets to confirm the robustness and generalizability of these findings. Additionally, given the highly imbalanced nature of the standard MRI cohort, future studies should assess the stability of performance across different dataset splits to better understand how model performance is influenced by variability in training data composition. Integrating interpretability features into DL models will also be critical to support clinical adoption and foster trust in these systems.

\appendix    
\label{sec:additional_results}
\section{Dataset Characteristics}
\label{app:dataset}
Out of the 586 MRIs, there were 109 (18.5\%) with Bankart lesions based on intraoperative photos in correlation with the surgeon's operative note (diagnostic gold standard). Of the 117 MRIs used to test the developed model, there were 22 (18.8\%) with Bankart lesions. This ratio was approximately the same in the training, validation, and testing sets due to using random stratified splitting. The MRA and standard MRI groups varied significantly. Of the 251 MRI arthrograms, 80 (31.9\%) had Bankart lesions, while only  8.6\% (29/335) of standard MRIs had Bankart lesions (p $<$ 0.001). MRA patients were significantly younger than standard MRI patients (48.6 versus 31.0 years old, p $<$ 0.001). Moreover, there was a lower percentage of female patients in the MRA group than the standard MRI group (35.1\% versus 50.1\%; p $<$ 0.001). The percent of right-sided exams between both groups was similar (61.0\% MRA versus 63.3\% standard MRI; p=0.565). However, there was a higher proportion of 3.0T scans for MRAs than for standard MRIs (73.3\% versus 54.6\%; p $<$ 0.001). 

\section{System Requirements}
 All model training and inference was performed on one NVIDIA A100-40GB GPU. Models were implemented using Python (version 3.10) and the PyTorch library (version 2.3.1). 

\acknowledgments 
 We would like to thank the UChicago Center for Research Informatics (CRI) High-Performance Computing (HPC) Cluster (Randi) staff for providing access to the HPC and secure storage resources used in this project. We would also like to thank the UChicago CRI Clinical Research Data Warehouse \& Human Imaging Research Office (HIRO) for their roles in data collection. 

 This project was supported by the National Center for Advancing Translational Sciences (NCATS) of the National Institutes of Health (NIH) through Grant Number UL1TR002389-07 that funds the Institute for Translational Medicine (ITM). The content is solely the responsibility of the authors and does not necessarily represent the official views of the NIH.

 M.S. was funded by the U.S. Department of Energy, Office of Science, Office of Advanced Scientific Computing Research, Department of Energy Computational Science Graduate Fellowship under Award Number DE-SC0023112.

\bibliography{references} 
\bibliographystyle{spiebib} 

\end{document}